\documentclass[letterpaper,10pt,conference]{ieeeconf}
\IEEEoverridecommandlockouts
\usepackage{amsmath,amsfonts,amssymb}
\usepackage{xcolor}
\usepackage{algorithmic}
\usepackage{algorithm}
\usepackage{array}
\usepackage{textcomp}
\usepackage{stfloats}
\usepackage{url}
\usepackage{graphicx}
\usepackage{cite}
\usepackage{import}
\usepackage{tabularx}
\usepackage{multirow}
\usepackage{colortbl}
\usepackage{booktabs}
\usepackage{siunitx}
\usepackage{subcaption}
\usepackage[colorlinks=true,citecolor=blue,linkcolor=blue,urlcolor=blue]{hyperref}
\usepackage[nameinlink,capitalise]{cleveref}
\usepackage{acronym}
\usepackage[table]{xcolor}
\definecolor{Black}{gray}{0.0}
\definecolor{Purple}{rgb}{0.77,0.12,0.64}
\def\method{WALT}
\title{WALT: Learning World-Model-Aligned Latent Trajectories \\ for Autonomous Driving}
\author{Mingkai Jia$^{1,2}$, Jiaxin Guo$^{3}$, Zhijian Shu$^{2,4}$, Jiawei Xu$^{2,5}$, Mingxiao Li$^{2,\dagger}$, Jintao Cheng$^{1}$, Ping Tan$^{1,*}$, Wei Yin$^{2}$
\thanks{$^{1}$ The
Hong Kong University of Science and Technology.}%
\thanks{$^{2}$ Horizon Robotics.}%
\thanks{$^{3}$ The Chinese University of Hong Kong.}%
\thanks{$^{4}$ Nanjing University of Posts and Telecommunications.}
\thanks{$^{5}$ Nankai University.}
\thanks{$\dagger$ Project lead: Mingxiao Li.}
\thanks{$*$ Corresponding author: Ping Tan (pingtan@ust.hk).}
}

\begin{document}

\makeatletter
\let\@oldmaketitle\@maketitle%
\renewcommand{\@maketitle}{%
    \@oldmaketitle%
    \centering
    \includegraphics[width=\linewidth]{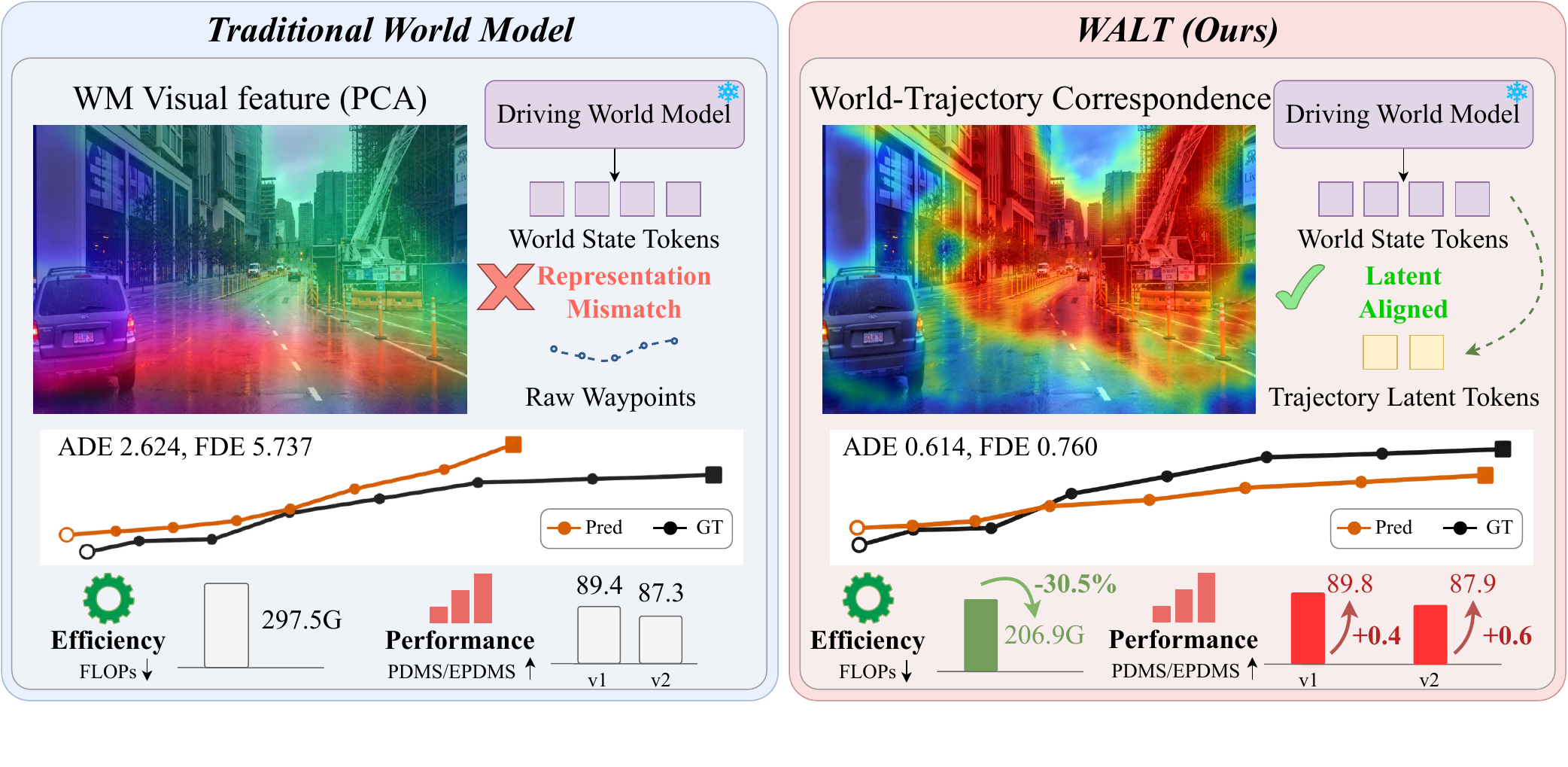}
    \vspace{-38pt}
    \captionof{figure}{%
    \small \textbf{Overview of \method{}.} 
\textbf{Left:} The baseline directly generates raw waypoints from frozen world-state tokens, leaving a representation mismatch between rich visual features and geometric trajectory coordinates. The PCA overlay visualizes the world model's rich feature structure.
\textbf{Right:} \method{} addresses this mismatch by learning compact trajectory latents aligned with the frozen world-model features. The world--trajectory correspondence map visualizes cosine similarity between the aligned trajectory and visual features, with red indicating higher similarity.
This aligned latent interface improves planning scores on NAVSIMv1 and NAVSIMv2 while reducing FLOPs.}
    \vspace{-17pt}

    \label{fig:teaser}
    \bigskip
}
\makeatother

\maketitle

\setcounter{figure}{1}

\begin{abstract}
Driving world models learn rich predictive representations of the surrounding environment from visual observations, yet accurate visual prediction does not necessarily translate into effective trajectory planning.
We argue that a key bottleneck lies in the mismatch between visual world states and raw geometric trajectories, which may limit the planner's ability to exploit action-relevant semantics encoded by the world model.
To address this issue, we propose World-Model Alignment for Latent Trajectories (\method{}), which learns a compact generative trajectory latent space by transferring information from a frozen pretrained driving world model without modifying the world model itself.
Rather than directly generating raw waypoints, \method{} maps them into compact representations through a dual-branch trajectory autoencoder and transfers semantic knowledge from the frozen visual world model into this trajectory space, encouraging the learned action representation to capture scene-level cues relevant to future motion and planning.
Beyond our proposed formulation, we systematically study latent learning based on Joint-Embedding Predictive Architectures (JEPA) and feature alignment following Representation Alignment (REPA) to investigate how trajectory-only representation learning affects downstream planning.
We evaluate \method{} on the NAVSIM benchmarks. Relative to the raw-waypoint baseline, \method{} improves PDMS from 89.4 to 89.8 on NAVSIMv1 and EPDMS from 87.3 to 87.9 on NAVSIMv2 while reducing trajectory planner FLOPs by 30.5\%.
These results suggest that preserving world representations while extracting action-relevant information provides an effective interface for world-model-based trajectory planning.
\end{abstract}

\section{Introduction}
Driving world models (DWMs) learn predictive representations from past observations and provide forward-looking information for autonomous driving~\cite{drivewm,driveworld,epona,world4drive}.
By forecasting future observations or latent states, they encode visual content and environmental dynamics for future prediction and motion planning.
Recent DWMs further strengthen these predictive states with supervision beyond visual prediction, including geometric targets, semantic features, and perception-derived signals~\cite{world4drive,epona,drivevla-w0,pwm,drivelaw,xu2025ad,cheng2024mf}.

However, high-fidelity visual prediction or a visually rich latent space does not ensure reliable trajectory planning.
Fig.~\ref{fig:teaser} illustrates how this mismatch can affect the action-side response and how \method{} addresses it.
The action head still has to map a rich predictive state to an action representation, and common planners expose this interface through raw waypoints, local motion increments, or discrete motion tokens~\cite{epona,drivingworld,drivinggpt,diffusiondrive,litevggt,chen2025focused,guo2026salon3r,guo2025endo3r,guo2022visual,jia2025mgvq,zhou2026thinklocallyrefineglobally}.
Recent unified visual--motion models address this gap by incorporating trajectory representations into world-model training and using the coupled representations for planning~\cite{drivelaw,gui2026bridging}.
Although this strategy enables visual prediction and motion planning to interact, it couples the action interface to a redesigned multimodal world model that jointly trains fused visual and motion representations.
This raises a question: \textit{Can we transfer a frozen pretrained DWM's visual knowledge into a trajectory generation space without modifying the world model?}

To answer this question, we propose World-Model Alignment for Latent Trajectories (\method{}), which to our knowledge is the first framework to learn a compact generative trajectory latent space by transferring information from a frozen pretrained driving world model without modifying the world model itself.
\method{} encodes raw waypoints with a dual-branch trajectory autoencoder.
The reconstruction branch preserves the geometric information required to recover raw waypoints, while the semantic branch transfers knowledge from the frozen visual world-model representation to capture scene-level cues relevant to future motion and planning.
The resulting latent serves as the trajectory generation space and is decoded back into planner waypoints.

Beyond the proposed alignment, we design and systematically study Joint-Embedding Predictive Architectures (JEPA) latent learning~\cite{maes2026leworldmodel}  and Representation Alignment (REPA) feature alignment~\cite{yu2024representation} to investigate how trajectory-only representation learning affects downstream planning.
Trajectory-only supervision can capture intrinsic motion structure, but it does not explicitly relate that structure to the surrounding visual scene.
The limited planning gains of these methods in our experiments support our use of world-model alignment to provide scene-derived supervision beyond trajectory geometry.
Experiments on NAVSIMv1~\cite{navsim-v1} and NAVSIMv2~\cite{navsim-v2} show that \method{} improves planning performance and reduces trajectory-generation FLOPs relative to the raw-waypoint baseline.

\begin{figure*}[t]
    \centering
    \includegraphics[width=0.96\linewidth]{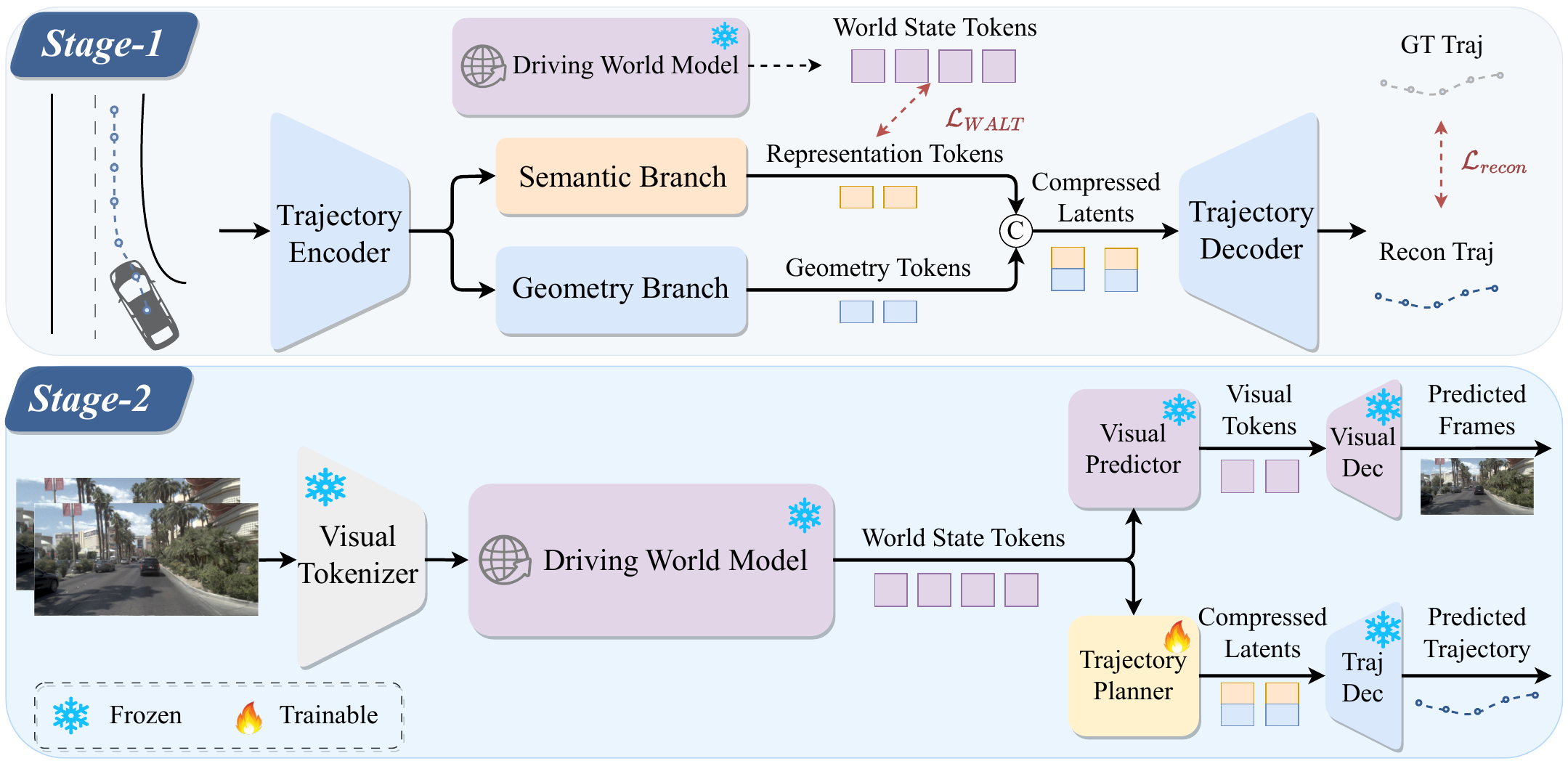}
    \caption{\textbf{\method{} pipeline.} In Stage 1, we train the dual-branch trajectory autoencoder with a frozen world model. The semantic branch feature is aligned with world state tokens with $\mathcal{L}_\text{WALT}$, while the geometric branch preserves reconstruction details and jointly trained with $\mathcal{L}_{recon}$. In Stage 2, we freeze the tokenizer and train the trajectory planner head of the frozen world model to generate latents, which are decoded into waypoints for future trajectory prediction.}
    \label{fig:pipeline}
    \vspace{-15pt}
\end{figure*}

Our contributions are summarized as follows:
\begin{itemize}
    \item We propose \method{}, which to our knowledge is the first framework that learns a compact generative trajectory latent space by transferring information from a frozen driving world model without modifying the world model itself.
    \item We construct a dual-branch trajectory autoencoder that preserves metric geometry while transferring semantic knowledge from the frozen world-model representation, and systematically study JEPA-style and REPA-style trajectory-only designs.
    \item We validate the effectiveness and efficiency of our approach on NAVSIMv1~\cite{navsim-v1} and NAVSIMv2~\cite{navsim-v2}, showing a 0.4 PDMS gain, 0.6 EPDMS improvement and a 30.5\% FLOPs reduction in trajectory-generation FLOPs over the waypoint baseline.
\end{itemize}

\section{Related Work}
\subsection{Driving World Models}
Driving world models learn predictive states from observation histories by forecasting future sensory observations or latent features~\cite{tu2025survey,kong20253d,drivewm,driveworld,world4drive}.
Video-space models use future image generation to capture appearance changes and temporal dynamics, while structured and latent models predict compact scene features that retain spatial relations without requiring pixel-level rollout~\cite{law,resworld,li2025imagidrive,yang2025worldrft}.
Although their prediction targets differ, these approaches progressively enrich the hidden state with information about scene layout, object motion, and temporal context.
This shift from output fidelity alone toward reusable predictive representations has made the latent state an increasingly important interface for downstream driving tasks~\cite{drivevla-w0,pwm,drivelaw}.
Beyond RGB forecasting, recent work introduces geometric or semantic signals to strengthen the predictive representation.
Epona~\cite{epona} combines compact visual features with autoregressive future prediction in a diffusion world model.
EponaV2~\cite{eponav2} extends this representation by requiring the inferred future state to support future image, metric-depth, and foundation-model semantic prediction.
These complementary targets encourage a future representation that retains visual content together with explicit geometry and semantic context.
Our focus is not to redesign the predictive state, but to learn an action-side trajectory space that can make better use of the information already encoded within it.

\subsection{Trajectory Representation for Planning}
Action interfaces in autonomous driving range from low-level controls and adjacent-frame motion increments to continuous waypoint sequences and learned motion tokens.
DrivingWorld and DrivingGPT encode local ego motion such as changes in planar position and heading into autoregressive tokens~\cite{drivingworld,drivinggpt}.
Local increments provide a temporally regular prediction target and fit naturally within next-token models.
However, recovering a complete path requires accumulation over time, and discretized variants additionally introduce quantization while leaving the full-path geometry implicit.
Continuous planners instead predict the future path directly as a sequence of waypoints~\cite{epona,diffusiondrive,goalflow}.
The common current-ego-centered representation should be distinguished from both global absolute position and adjacent-frame relative motion.
It remains local and translation invariant, yet directly preserves the metric geometry and behavior of the full future path without repeated integration.
WorldDrive moves toward a learned motion interface by optimizing motion and vision representations through scene generation and transferring them to planning~\cite{gui2026bridging}.
This design couples motion structure with the visual model, while \method{} focus on an aligned representation under a frozen world model.
We encode the complete current-ego-centered future path into a compact continuous latent, preserve metric recovery through an explicit decoder, and separately organize its semantic component for alignment with the predictive world state.

\subsection{Representation Learning}
Joint-embedding predictive architectures learn by predicting target representations rather than reconstructing observations.
I-JEPA establishes this principle for images, V-JEPA extends it to video, and V-JEPA 2 demonstrates that self-supervised video representations can support understanding, prediction, and planning~\cite{assran2023self,bardes2024v,assran2025v}. 
LeWorldModel further combines next-embedding prediction with SIGReg to stabilize end-to-end latent world-model learning~\cite{maes2026leworldmodel}.
A related line studies the representations formed inside diffusion models and how stronger external features can improve generative learning~\cite{mi2025data,zhong2025anytalker,dinotok}.
REPA reverses the transfer direction by aligning diffusion-transformer hidden states with clean representations from a frozen self-supervised encoder~\cite{yu2024representation}.
In autonomous driving, Drive-JEPA adapts video JEPA pretraining to driving data and combines the predictive visual encoder with multimodal trajectory distillation~\cite{wang2026drive}.
Its primary learned representation remains visual and supports a proposal-centric planner.
Auto-JEPA more directly learns an action-oriented intent space by aligning a predicted context embedding with the latent representation of a future ego trajectory~\cite{yang2026auto}.
That latent acts as a retrieval key for a fixed trajectory memory at inference.
Unlike this retrieval-based formulation, \method{} treats the pretrained world model as a frozen semantic teacher and transfers its action-relevant knowledge into a separately learned generative trajectory space.
The learned latent is generated by the planner and explicitly decoded back into raw waypoints.
We also design and systematically investigate JEPA-inspired latent learning and REPA-style downstream alignment to study trajectory-only representations without world-model transfer.

\section{Method}
\subsection{Overview}
The \method{} pipeline is illustrated in Fig.~\ref{fig:pipeline}. In Stage 1, we train a dual-branch trajectory autoencoder to compress raw waypoints into a compact latent representation (Sec.~\ref{sec:tokenizer}). The training objective includes a reconstruction loss and a world-model alignment loss that transfers information from the frozen world model into the trajectory latent space (Sec.~\ref{sec:walt}). In Stage 2, we employ the aligned trajectory latent as the generation space for the action head of the frozen world model (Sec.~\ref{sec:latent_generation}). By generating in this aligned latent space, the planner aims to better exploit the semantic information encoded in the world model. A systematic study of trajectory-only representation learning is also conducted to evaluate the impact of visual world-model alignment on downstream planning (Sec.~\ref{sec:trajectory_analysis}).

\subsection{Dual-Branch Trajectory Tokenizer}
\label{sec:tokenizer}

Trajectory representation for planning is crucial but not unified among existing works~\cite{epona,drivingworld,drivinggpt,diffusiondrive}.
Directly modeling raw waypoints is simple and preserves metric geometry, but it does not explicitly separate semantic information from geometric information.
Therefore, we design a dual-branch trajectory autoencoder that compresses raw waypoints into a compact latent representation with separate semantic and reconstruction components.
Following previous work~\cite{eponav2}, we consider a sequence of future trajectory points $A=\{A_i\}_{i=N+1}^{N+P}$, where $N$ denotes the final observed time index and $P$ is the prediction horizon, and compress it into $K$ latent tokens with a downsampling ratio $q=P/K$.
The two branches share an encoder and use separate refiners to produce semantic features $z_{sem}$ and reconstruction features $z_{rec}$.
These features are concatenated along the channel dimension as $z_A=[z_{sem};z_{rec}]$, where $[\cdot;\cdot]$ denotes concatenation.
A decoder then maps the concatenated latent $z_A$ back to the original trajectory space as $\hat{A}=\mathcal{D}(z_A)$.
The training objective includes an $\ell_1$ reconstruction loss $\mathcal{L}_\text{rec}=\mathbb{E}_{A}[\lVert\hat{A}-A\rVert_1]$ and a world-model alignment loss that transfers information from the frozen world model into the trajectory latent space:
\begin{equation}
    \mathcal{L}_\text{tokenizer}
    = \lambda_{rec}\mathcal{L}_\text{rec}
    + \lambda_{align}\mathcal{L}_\text{WALT}.
\end{equation}
The world model remains frozen throughout tokenizer training.
The dual-branch design encourages the semantic component to capture action-relevant information while the reconstruction component preserves the geometric structure of the trajectory.
To encourage $z_{rec}$ to preserve sufficient information for trajectory reconstruction, we mask the entire semantic component $z_{sem}$ with probability $p_{sem}$ while retaining $z_{rec}$. This encourages the reconstruction branch to retain sufficient geometric information for accurate trajectory recovery even when semantic information is unavailable.
The resulting latent serves as the trajectory generation space and is decoded back into planner waypoints.

\begin{table*}[!t]
\caption{\textbf{Reported results on NAVSIMv1~\cite{navsim-v1}.} All metrics are on a 0--100 scale. Higher is better. Bold and underlined values indicate the best and second-best distinct scores among learned methods, respectively. These results include different model and training configurations.}
\label{tab:navtest-v1}
\centering
\setlength{\tabcolsep}{4mm}
\resizebox{0.9\linewidth}{!}{%
\begin{tabular}{l|c|ccccc>{\columncolor{gray!15}}c}
\toprule
Method & Venue & NC & DAC & EP & TTC & C & PDMS \\
\midrule
Human & -- & 100 & 100 & 87.5 & 100 & 99.9 & 94.8 \\
\midrule
LAW~\cite{law} & ICLR'25 & 96.4 & 95.4 & 81.7 & 88.7 & \underline{99.9} & 84.6 \\
DrivingGPT~\cite{drivinggpt} & ICCV'25 & 98.9 & 90.7 & 79.7 & 94.9 & 95.6 & 82.4 \\
World4Drive~\cite{world4drive} & ICCV'25 & 97.4 & 94.3 & 79.9 & 92.8 & \textbf{100} & 85.1 \\
Epona~\cite{epona} & ICCV'25 & 97.9 & 95.1 & 80.4 & 93.8 & \underline{99.9} & 86.2 \\
PWM~\cite{pwm} & NeurIPS'25 & 98.6 & 95.9 & 81.8 & 95.4 & \textbf{100} & 88.1 \\
TISA~\cite{zhao2025autoregressive} (w/o MOPT)& ICRA'26 &98.0&95.5&81.1&93.8&--&86.8 \\
AdaThinkDrive~\cite{luo2025adathinkdrive} (w/o RL) & ICRA'26 & 98.9 & 95.3 & 80.6 & 96.0 & \textbf{100} & 87.5 \\
Mimir~\cite{xing2025mimir} & RA-L'26 &98.2 & \textbf{97.5} & \textbf{83.6} & 94.6 & \textbf{100} & 89.3 \\
PRIX~\cite{wozniak2026prix} & RA-L'26 & 98.1 & 96.3 & \underline{82.3} & 94.1 & \textbf{100} & 87.8 \\
ARTEMIS~\cite{feng2025artemis} & RA-L'26 & 98.3 & 95.1 & 81.4 & 94.3 &\textbf{100} & 87.0 \\
DriveVLA-W0~\cite{drivevla-w0} & ICLR'26 & 98.4 & 95.3 & 80.9 & 95.2 & \textbf{100} & 87.2 \\
DriveLaW~\cite{drivelaw} & CVPR'26 & \underline{99.0} & 97.1 & 81.3 & \textbf{96.7} & \textbf{100} & 89.1 \\
EponaV2~\cite{eponav2} (w/o RL) & arXiv'26 & 98.6 & \underline{97.3} & \textbf{83.6} & 95.3 & \underline{99.9} & \underline{89.4}\\
\textbf{\method{} (Ours)} & -- & \textbf{99.1} & 97.0 & \textbf{83.6} & \underline{96.3} & \textbf{100} & \textbf{89.8} \\
\bottomrule
\end{tabular}}
\vspace{-10pt}
\end{table*}

\subsection{World-Model Alignment for Latent Trajectories}
\label{sec:walt}

To exploit the rich predictive representation learned by the frozen world model, we aim to transfer its action-relevant semantics into the trajectory latent space.
To this end, \method{} aligns the semantic branch with the world model's visual features without modifying the world model itself.
We use the frozen world-model feature as a contextual teacher, keeps the visual encoder and backbone remain frozen, and the task-specific prediction heads are not used during tokenizer training.
For each scene frame and its associated trajectory, the teacher produces hidden states $H_w\in\mathbb{R}^{N_b\times M\times d_w}$, where $N_b$ denotes the flattened batch and frame dimension, $M$ is the number of teacher tokens, and $d_w$ is their channel dimension.
A learnable projector maps the student semantic tokens to the teacher channel dimension, yielding $U=g(z_{sem})$.
We formulate a CLIP-style contrastive objective. For student sample $i$ and teacher sample $j$, the sample-level score averages all $KM$ token-pair cosine similarities:
\begin{equation}
    S_{ij}=\frac{1}{KM}\sum_{k=1}^{K}\sum_{\ell=1}^{M}
    \frac{U_{ik}^{\mathsf T}H_{w,j\ell}}
    {\lVert U_{ik}\rVert_2\lVert H_{w,j\ell}\rVert_2}.
\end{equation}
Positive pairs share the same sample and frame index, while every other gathered sample or frame is treated as a negative.
Student features are gathered across GPUs with gradients and frozen teacher features are gathered without gradients, so the candidate set spans the global batch.
Let $y_i=i$ denote the matching teacher index for student sample $i$.
Following the symmetric CLIP objective~\cite{radford2021learning}, we use
\begin{equation}
    \mathcal{L}_\text{WALT}=\frac{1}{2}\operatorname{CE}(\alpha S,y)
    +\frac{1}{2}\operatorname{CE}(\alpha S^{\mathsf T},y).
\end{equation}
The learnable logit scale $\alpha=\exp(s)$ controls the concentration of the contrastive distribution and is bounded during optimization.
Through this objective, \method{} transfers world-model information into the action representation during tokenizer training without concatenating scene features into the trajectory latent.

\subsection{Latent Trajectory Generation}
\label{sec:latent_generation}
After tokenizer training, we freeze the trajectory tokenizer and retain the original frozen world-model pathway.
We use a rectified-flow planner as the trajectory head following previous works~\cite{flow-ode,flow-matching,rectified-flow,svg,svg-t2i,eponav2}.
The head is trained with a conditional flow-matching objective, but its target is changed from $P$ raw waypoint tokens to $K$ concatenated latent tokens $z_A$.
That is, the flow-matching trajectory head predicts the latent trajectory velocity $\hat{v}_A$ in the learned compact representation space rather than the velocity of raw waypoints.

Given $\epsilon\sim\mathcal{N}(\mathbf{0},\mathbf{I})$ and $t\sim\mathcal{U}[0,1]$, we construct
\begin{equation}
    z_t=(1-t)\epsilon+t z_A,
    \qquad v_{target}=z_A-\epsilon.
\end{equation}
Conditioned on the frozen world-model state $H_w$, the trajectory head minimizes
\begin{equation}
    \mathcal{L}_\text{flow}
    =\mathbb{E}_{A,\epsilon,t}
    \left[
    \left\lVert
    \hat{v}_A(z_t,t;H_w)-(z_A-\epsilon)
    \right\rVert_2^2
    \right].
\end{equation}
At inference, we initialize $\hat{z}_0\sim\mathcal{N}(\mathbf{0},\mathbf{I})$ and integrate the predicted velocity from $t=0$ to $t=1$:
\begin{equation}
    \hat{z}_{t+\Delta t}
    =\hat{z}_t+\Delta t\,\hat{v}_A(\hat{z}_t,t;H_w).
\end{equation}
The decoder then maps the generated latent to the predicted waypoints as $\hat{A}=\mathcal{D}(\hat{z}_1)$.
With the learned trajectory latent space, the flow-matching planner generates compact trajectory representations aligned with the world model, supporting effective planning with reduced trajectory head computation.

\begin{table*}[!t]
\caption{\textbf{Comparison on NAVSIMv2~\cite{navsim-v2}.} All metrics are on a 0--100 scale, where higher is better. $^*$ denotes results reported using the original evaluator before the human-penalty aggregation update. Bold and underlined values indicate the best and second-best distinct scores among unstarred learned methods.}
\label{tab:navtest-v2}
\centering
\setlength{\tabcolsep}{3mm}
\resizebox{0.9\linewidth}{!}{%
\begin{tabular}{lccccccccc>{\columncolor{gray!15}}c}
\toprule
Method  & NC & DAC & DDC & TLC & EP & TTC & LK & HC & EC & EPDMS \\
\midrule
Human  & 100 & 100 & 99.8 & 100 & 87.4 & 100 & 100 & 98.1 & 90.1 & 94.5 \\
\midrule
DriveVLA-W0~\cite{drivevla-w0} & 98.4 & 95.2 & 99.4 & \textbf{99.9} & 86.6 & 97.9 & {97.8} & \textbf{98.3} & \underline{82.7} & 86.9 \\
ARTEMIS$^*$~\cite{feng2025artemis} & \textbf{98.3} & 95.1 & 98.6 &99.8 & 81.5 & 97.4 & 96.5 & 98.3 & -- & 83.1\\
PRIX$^*$~\cite{wozniak2026prix} & 98.0 & 85.6 & \underline{99.5} & 99.8 & \underline{87.4} & 97.2 & 97.1 & \textbf{98.3} & \textbf{87.6} & 84.2 \\
DriveWorld-VLA~\cite{liu2026driveworldvla}&\textbf{98.6}&\textbf{99.1}&\textbf{99.6}&99.8&\underline{87.4}&97.9&97.0&97.8&78.6&86.8 \\

EponaV2~\cite{eponav2} (w/o RL) & 98.4 & 96.7 & \textbf{99.6} & \textbf{99.9} & \textbf{87.6} & \underline{98.1} & \textbf{98.0} & \underline{98.1} & 68.3 & \underline{87.3} \\
\textbf{\method{} (Ours)} & \underline{98.5} & \underline{96.8} & \textbf{99.6} & \textbf{99.9} & \underline{87.4} & \textbf{98.2} &\underline{97.9}&\textbf{98.3} & 73.4 &\textbf{87.9}\\

\bottomrule
\end{tabular}}
\vspace{-10pt}
\end{table*}

\subsection{Trajectory-Only Representation Analysis}
\label{sec:trajectory_analysis}
To assess the contribution of visual world-model alignment, we further investigate two trajectory-only representation learning methods, JEPA-Traj and REPA-Traj, under the same planner setting.
JEPA-Traj uses latent trajectory generation, whereas REPA-Traj retains raw-waypoint generation and adds auxiliary feature alignment.
Joint-embedding predictive architectures (JEPA) learn by predicting target representations and are commonly used in self-supervised visual representation learning~\cite{assran2023self,bardes2024v,assran2025v}.
To apply JEPA to the trajectory modality, we treat continuous waypoints as sequential inputs, extract overlapping sub-trajectories, and encode them with the same semantic tokenizer.
Let $z_{sem}^{(1)}$ and $z_{sem}^{(2)}$ be the semantic tokens of two overlapping sub-trajectories, where $z_{sem}^{(1)}$ corresponds to the earlier sub-trajectory and $z_{sem}^{(2)}$ to the later one.
Conditioned on the time offset at which the overlap begins, a predictor is trained to predict the later sub-trajectory's semantic tokens from the earlier ones.
To prevent representation collapse, we use SIGReg~\cite{maes2026leworldmodel} regularization.
The representation-learning objective is
\begin{equation}
    \mathcal{L}_\text{JEPA}
    =\lambda_\text{sigreg}\mathcal{L}_\text{sigreg}
    +\lambda_\text{pred}\mathcal{L}_\text{pred},
\end{equation}
where $\mathcal{L}_\text{pred}$ is the $\ell_2$ prediction loss between the predicted and target semantic tokens, $\mathcal{L}_\text{sigreg}$ is the SIGReg regularization loss, $\lambda_\text{sigreg}$ and $\lambda_\text{pred}$ are loss weights.
This objective replaces the world-model alignment term during tokenizer training, while the reconstruction loss is retained.
Visualizations of within-class and between-class similarity distributions, class-pair heatmaps, and t-SNE embeddings in Fig.~\ref{fig:trajectory_representation} show that JEPA-style representation learning helps separate different driving behavior classes.

Beyond using JEPA-style trajectory-only representation learning for latent generation, we also investigate REPA-style~\cite{yu2024representation} feature alignment, which uses the learned trajectory representation in a different way.
REPA~\cite{yu2024representation} uses visual foundation-model representations as teacher targets for intermediate diffusion-transformer features.
We apply this idea to trajectory-only representation learning, using the learned trajectory representation as a teacher target for the trajectory stream of the flow-matching planner.
Let $h_{traj}$ be the trajectory-stream representation before the joint single-stream blocks, and let $z_{sem}^{\ast}=\operatorname{sg}(z_{sem})$ be the frozen semantic target from the JEPA-style trained trajectory autoencoder, where $\operatorname{sg}$ denotes stop-gradient.
A projector $g_R$ maps trajectory sequence channels to the semantic target dimension.
An auxiliary cosine-distance loss between $g_R(h_{traj})$ and $z_{sem}^{\ast}$ is added to the flow-matching objective.
This REPA-style feature alignment keeps raw waypoints as the flow-matching target, allowing us to examine whether the learned trajectory representation helps the downstream trajectory stream generate better waypoints.
The results in Table~\ref{tab:main_results} show only marginal PDMS gains for JEPA-Traj and REPA-Traj over the raw-waypoint baseline.
These results motivate incorporating visual world-model supervision into trajectory representation learning.

\begin{figure*}[t]
    \centering
    \includegraphics[width=0.95\linewidth]{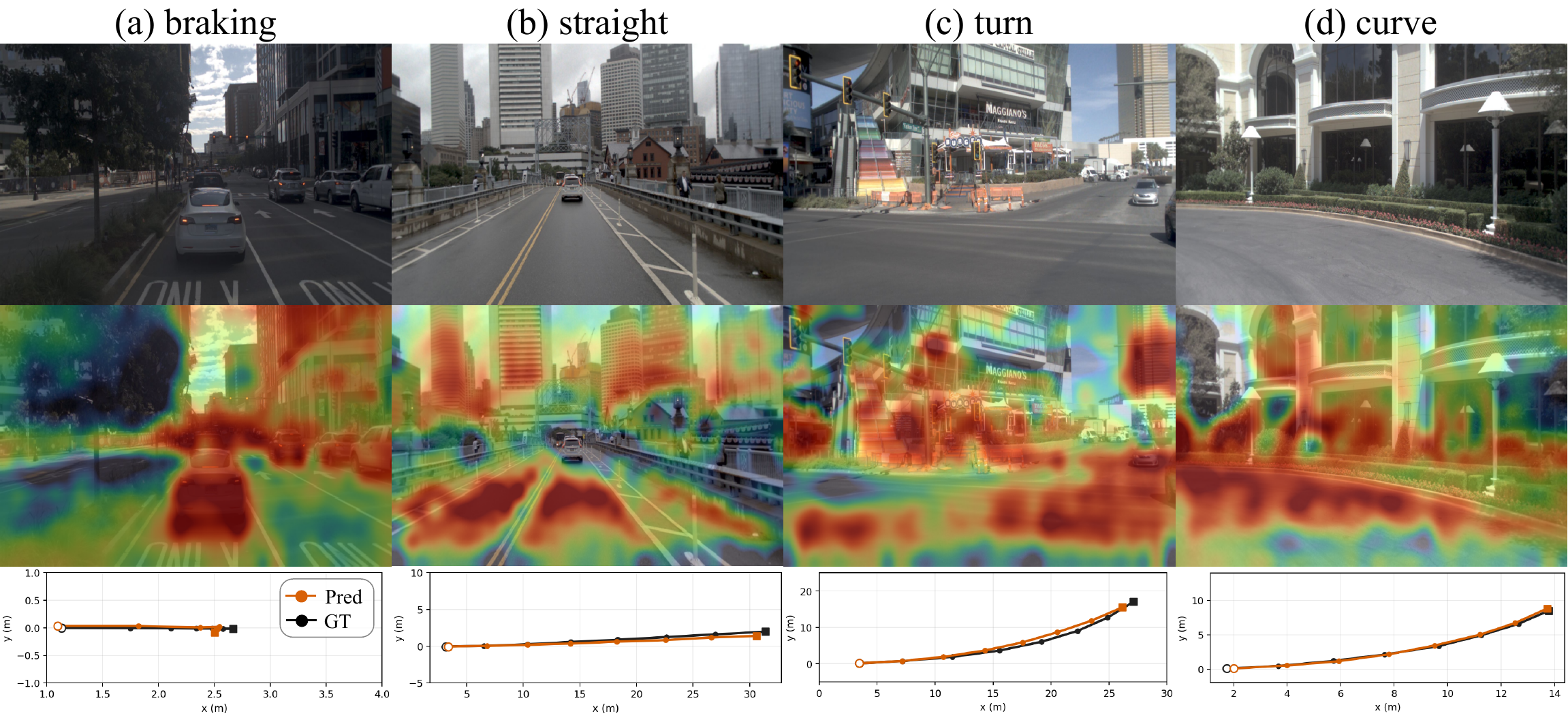}
    \caption{\textbf{Qualitative trajectories and world-model alignment.}
Columns show cases for braking, straight driving, turn, and curve.
From top to bottom: front-view images, world--trajectory correspondence maps, and predicted trajectories (orange) with ground truth (black).
The maps visualize cosine similarity between projected semantic tokens from planner-generated latents and frozen visual world-model tokens, with warmer colors indicating higher similarity.}
    \label{fig:qualitative}
    \vspace{-10pt}
\end{figure*}
\section{Experiments}

\subsection{Experimental Setup}

\noindent\textbf{Benchmarks and metrics.}
We evaluate on NAVSIMv1 navtest using the official planning protocol~\cite{navsim-v1}. We report PDMS together with no-at-fault collision (NC), drivable-area compliance (DAC), ego progress (EP), time-to-collision (TTC), and comfort (C). We additionally evaluate on NAVSIMv2~\cite{navsim-v2}, reporting EPDMS and its associated component metrics, including traffic light compliance~(TLC) and lane keeping~(LK), while expanding comfort~(C) into history comfort~(HC) and extended comfort~(EC). For all metrics reported, higher values indicating better performance. Evaluator differences about human filtering for reported NAVSIMv2 results are indicated in Table~\ref{tab:navtest-v2}.

\noindent\textbf{Implementation details.}
We instantiate \method{} with EponaV2~\cite{eponav2} as the pretrained driving world model, retaining its default model and planner-training settings except for the trajectory representation and associated alignment modules. We use the official benchmark data splits and evaluation protocols. Each trajectory contains eight ego-centric $(x,y,\mathrm{yaw})$ waypoints sampled at 0.5-s intervals. In Stage 1, the dual-branch tokenizer compresses the trajectory into two 32-dimensional tokens, each comprising 24 semantic and 8 reconstruction channels. We set the semantic masking probability to 0.5 and the alignment weight $\lambda_{align}$ to 0.1, and train the tokenizer for 100 epochs on 32 H20 GPUs with a fixed learning rate of $1\times 10^{-4}$, keeping the world-model backbone frozen. In Stage 2, we further freeze the trajectory autoencoder and fine-tune only the trajectory head.

\noindent\textbf{Representation variants.}
We compare five variants using the same frozen world-model backbone and evaluation protocol. 
JEPA-Traj adds trajectory-only latent prediction with SIGReg to tokenizer training, whereas REPA-Traj retains raw-waypoint generation and aligns intermediate planner features with the frozen JEPA-Traj semantic representation. Thus, these variants share the world-model conditioning source but differ in their generation targets or auxiliary supervision. For JEPA-Traj, SIGReg uses a weight of $5\times 10^{-4}$ with other settings same to LeWM~\cite{maes2026leworldmodel}. For REPA-Traj, cosine alignment is applied before the joint blocks in trajectory planning head with a weight of $0.1$. \method{} instead learns the trajectory latent through world-model alignment.

\subsection{Planning Performance}

\noindent\textbf{Benchmark comparison.}
To assess the planning benefits of our aligned trajectory representation without additional post-training, \method{} uses no reinforcement learning or similar refinement procedures, and we compare against results reported without such procedures where available, as indicated in the tables.
Table~\ref{tab:navtest-v1} compares \method{} with reported NAVSIMv1 results. 
\method{} achieves the best overall PDMS of 89.8, while also attaining the highest NC score and matching the best EP and comfort scores.
These results demonstrate a strong balance across progress, safety, and comfort.
Notably, these gains are obtained by learning a compact world-model-aligned trajectory space while keeping the pretrained world model unchanged, suggesting that this aligned latent can better exploit existing world knowledge for planning.

A consistent trend is observed on NAVSIMv2. As shown in Table~\ref{tab:navtest-v2}, \method{} achieves strongest EPDMS among all the methods. Compared to EponaV2 without RL, our scores improve in NC, DAC, TTC, HC, with the largest component gain in EC, which increases from 68.3 to 73.4. Meanwhile, DDC and TLC remain unchanged and match the best reported scores, while EP and LK decrease slightly by 0.2 and 0.1 points, respectively. Overall, the gains span several safety and comfort metrics while largely preserving progress and lane keeping, supporting the effectiveness of world-model-aligned trajectory representations across both benchmarks. 


\noindent\textbf{Effect of trajectory representation.}
Table~\ref{tab:main_results} isolates the effect of the action-side representation under the same frozen world-model backbone. The reconstruction-only autoencoder without semantic alignment changes PDMS from 89.42 to 89.48, indicating that the reconstruction-only encoded latent preserves information relevant to downstream planning. JEPA-Traj and REPA-Traj provide only marginal gains, suggesting that trajectory-only representation learning is insufficient to substantially improve the planner.

In contrast, \method{} reaches the best overall PDMS of 89.83, improving over both raw waypoints and the reconstruction-only autoencoder. 
Relative to raw waypoints, NC increases from 98.62 to 99.11 and TTC from 95.26 to 96.34, while maintaining competitive performance on the remaining metrics. 
Overall, the comparison highlights the importance of injecting world-aware semantics into the action representation, rather than learning the trajectory space purely from trajectory reconstruction or self-supervision.


\subsection{Trajectory Representation Analysis}

\begin{figure}[t]
    \centering
    \includegraphics[width=1.0\linewidth]{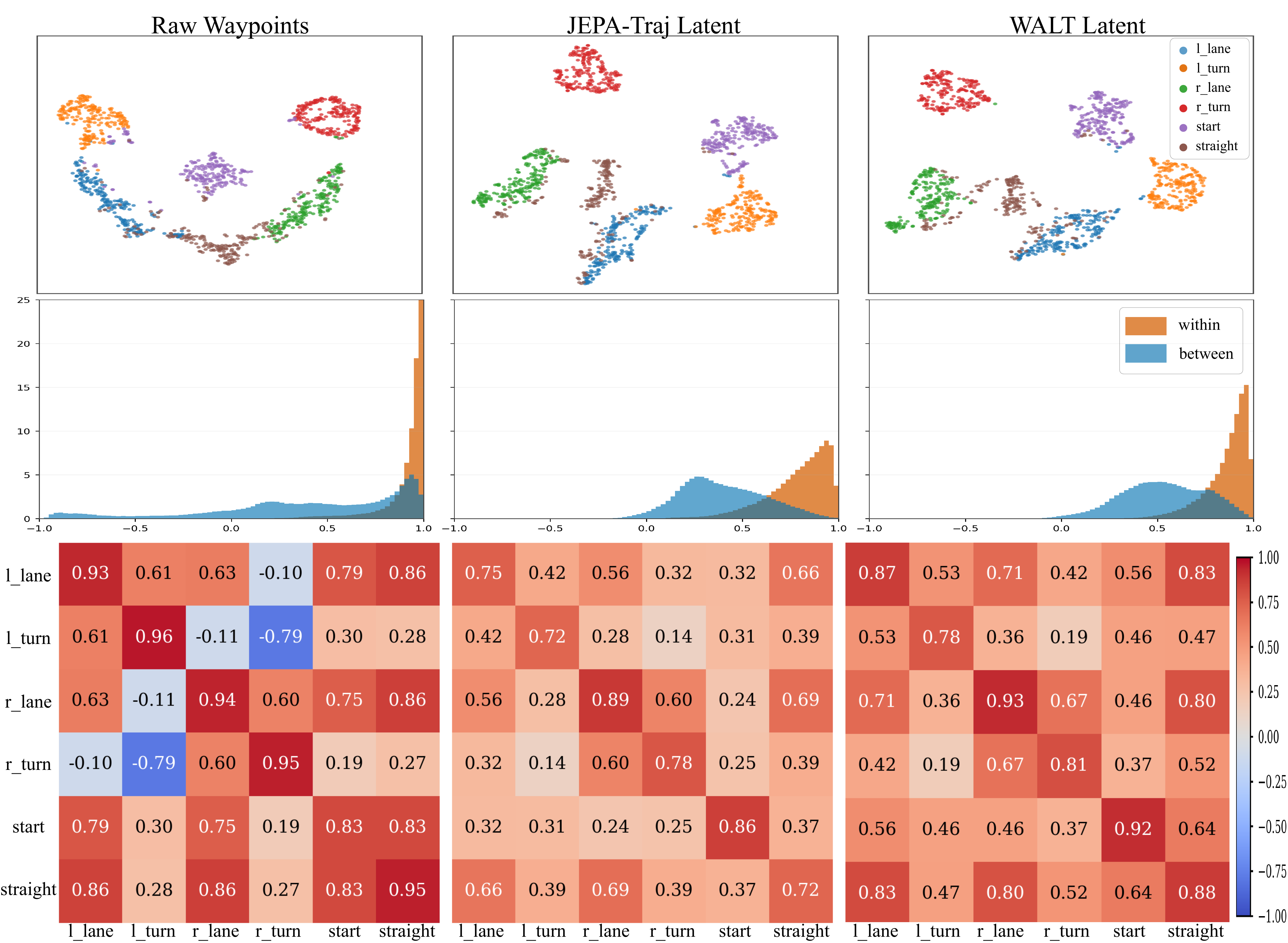}
    \caption{\textbf{Trajectory representation analysis on NAVSIMv1.} Columns compare raw waypoints, JEPA-Traj semantic latents, and \method{} semantic latents on identical samples. From top to bottom: t-SNE embeddings, within-class and between-class cosine-similarity distributions, and class-pair mean-similarity matrices. Learned representations use only their semantic components.}
    \label{fig:trajectory_representation}
    \vspace{-15pt}
\end{figure}

Fig.~\ref{fig:trajectory_representation} compares raw waypoints, JEPA-Traj semantic latents, and \method{} semantic latents on identical NAVSIMv1 test samples. We use six behavior classes: left lane change, left turn, right lane change, right turn, start, and normal straight. Each representation is flattened and $\ell_2$-normalized before computing pairwise cosine similarities. We examine within-class and between-class distributions, class-pair mean similarities, and t-SNE embeddings with cosine distance. Within-class distributions exclude self-pairs.

Raw waypoints exhibit strong within-class similarity but also substantial similarity between lane changes and straight driving, reflecting shared path geometry. JEPA-Traj reduces several cross-class similarities and produces more distinct behavior groups in the t-SNE visualization. Compared with JEPA-Traj, \method{} has higher within-class mean similarity for all six classes, but also retains higher similarity between several different classes. Thus, world-model alignment does not simply maximize separation between behavior labels. The planning results complement this observation: stronger separation under trajectory-only learning does not necessarily yield a larger PDMS gain. These visualizations characterize representation organization rather than establish which contextual information causes the planning improvement.

\subsection{Qualitative Results and Alignment Visualization}



Fig.~\ref{fig:qualitative} shows front-view observations, world-trajectory correspondence maps, and predicted trajectories for braking, straight driving, turning, and curve following. The predicted trajectories, decoded from planner-generated latents, closely follow the ground-truth paths across these examples. To visualize alignment, we compute cosine similarity between the projected semantic trajectory tokens and the front-view world-model visual tokens. The resulting scores are averaged over trajectory tokens and reshaped to the visual-token grid.

The correspondence maps reveal behavior-dependent spatial responses. During braking, high similarity concentrates on the leading vehicle and nearby roadway, while straight driving emphasizes the drivable road and lane direction. In turning and curved-road cases, responses shift toward road geometry and regions along the intended path. Some activation also appears on contextual background regions, indicating that these maps reflect representation-level correspondence rather than precise localization. Overall, these examples suggest that world-model alignment encourages trajectory latents to capture visual scene semantics relevant to each driving behavior.

\subsection{Computational Efficiency}

Under the NAVSIMv1 evaluation, the action head processes two latent trajectory tokens instead of eight raw waypoint tokens. Table~\ref{tab:efficiency} reports one denoising step 297.47 GFLOPs for  raw-waypoint and 206.88 GFLOPs for latent with decoding included, a 30.5\% reduction. The decoder contributes only a small fraction of the cost, and the frozen world-model computation is excluded. For an $N$-step rollout, the total trajectory-generation cost is $N F_{step}+F_{dec}$, where $F_{step}$ is the denoising-step cost and $F_{dec}$ is the one-time decoder cost. Since $F_{dec}$ is small, the full-rollout cost is approximately $N$ times the tabulated value.
\begin{table}[t]
\centering
\caption{\textbf{Action-side representation comparison on NAVSIMv1 with the same frozen world-model backbone.} Variants differ in generation target and representation supervision. All metrics are on a 0--100 scale. Higher is better.}
\label{tab:main_results}
\scriptsize
\setlength{\tabcolsep}{4pt}
\resizebox{\columnwidth}{!}{%
\begin{tabular}{@{}lcccccc@{}}
\toprule
Method & NC & DAC & EP & TTC & C & PDMS \\
\midrule
Raw waypoints baseline & 98.62 & 97.32 & 83.60 & 95.26 & 99.93 & 89.42 \\
$+$ trajectory AE (w/o $\mathcal{L}_\text{WALT}$) & 98.62 & 97.38 & 83.63 & 95.32 & 99.95 & 89.48 \\
JEPA-Traj & 98.55 & 97.37 & 82.85 & 95.91 & \textbf{100.00} & 89.46 \\
REPA-Traj & 98.48 & \textbf{97.46} & \textbf{83.79} & 95.11 & 99.98 & 89.49 \\
\method{} & \textbf{99.11} & 96.99 & 83.63 & \textbf{96.34} & \textbf{100.00} & \textbf{89.83} \\
\bottomrule
\end{tabular}}
\vspace{-5pt}
\end{table}

\begin{table}[t]
\centering
\caption{\textbf{Trajectory-generation cost for one denoising step under matched world-model conditioning.} The latent-generation cost includes trajectory decoding, and the frozen world-model computation is excluded.}
\label{tab:efficiency}
\resizebox{0.95\columnwidth}{!}{%
\begin{tabular}{lcc}
\toprule
Representation & Compression Ratio & GFLOPs $\downarrow$ \\
\midrule
Raw waypoint generation &  -- & 297.47 \\
\method{} latent generation  & 4$\times$ & 206.88 \\
\bottomrule
\end{tabular}
}
\vspace{-10pt}
\end{table}

\section{Conclusion}

We presented \method{}, which learns compact trajectory latents through alignment with representations from a frozen pretrained driving world model. Its dual-branch tokenizer combines metric reconstruction with world-model supervision, providing a latent space for trajectory generation without modifying the world-model backbone. Experiments on NAVSIMv1 and NAVSIMv2 show improved planning scores and reduction of FLOPs. Comparisons with reconstruction-only and trajectory-only alternatives support the value of world-model alignment in the evaluated setting. These findings motivate further study of learning world-model-aligned trajectory representations for autonomous driving.

\bibliographystyle{IEEEtran}
\bibliography{main}

@inproceedings{litevggt,
  title={Litevggt: Boosting vanilla vggt via geometry-aware cached token merging},
  author={Shu, Zhijian and Lin, Cheng and Xie, Tao and Yin, Wei and Li, Ben and Pu, Zhiyuan and Li, Weize and Yao, Yao and Cao, Xun and Guo, Xiaoyang and others},
  booktitle={Proceedings of the IEEE/CVF Conference on Computer Vision and Pattern Recognition},
  pages={36422--36432},
  year={2026}
}

@inproceedings{drivinggpt,
  title={{DrivingGPT}: {Unifying} driving world modeling and planning with multi-modal autoregressive transformers},
  author={Chen, Yuntao and Wang, Yuqi and Zhang, Zhaoxiang},
  booktitle={Proceedings of the IEEE/CVF International Conference on Computer Vision},
  pages={26890--26900},
  year={2025}
}

@inproceedings{diffusiondrive,
  title={{DiffusionDrive}: {Truncated} diffusion model for end-to-end autonomous driving},
  author={Liao, Bencheng and Chen, Shaoyu and Yin, Haoran and Jiang, Bo and Wang, Cheng and Yan, Sixu and Zhang, Xinbang and Li, Xiangyu and Zhang, Ying and Zhang, Qian and others},
  booktitle={Proceedings of the Computer Vision and Pattern Recognition Conference},
  pages={12037--12047},
  year={2025}
}

@inproceedings{goalflow,
  title={{GoalFlow}: {Goal-driven} flow matching for multimodal trajectories generation in end-to-end autonomous driving},
  author={Xing, Zebin and Zhang, Xingyu and Hu, Yang and Jiang, Bo and He, Tong and Zhang, Qian and Long, Xiaoxiao and Yin, Wei},
  booktitle={Proceedings of the Computer Vision and Pattern Recognition Conference},
  pages={1602--1611},
  year={2025}
}

@InProceedings{world4drive,
    author    = {Zheng, Yupeng and Yang, Pengxuan and Xing, Zebin and Zhang, Qichao and Zheng, Yuhang and Gao, Yinfeng and Li, Pengfei and Zhang, Teng and Xia, Zhongpu and Jia, Peng and Lang, XianPeng and Zhao, Dongbin},
    title     = {{World4Drive}: {End}-to-End Autonomous Driving via Intention-aware Physical Latent World Model},
    booktitle = {Proceedings of the IEEE/CVF International Conference on Computer Vision (ICCV)},
    month     = {October},
    year      = {2025},
    pages     = {28632-28642}
}

@inproceedings{epona,
  title={Epona: Autoregressive diffusion world model for autonomous driving},
  author={Zhang, Kaiwen and Tang, Zhenyu and Hu, Xiaotao and Pan, Xingang and Guo, Xiaoyang and Liu, Yuan and Huang, Jingwei and Yuan, Li and Zhang, Qian and Long, Xiao-Xiao and others},
  booktitle={2025 IEEE/CVF International Conference on Computer Vision (ICCV)},
  pages={27220--27230},
  year={2025},
  organization={IEEE}
}

@inproceedings{resworld,
title={{ResWorld}: {Temporal} Residual World Model for End-to-End Autonomous Driving},
author={Jinqing Zhang and Zehua Fu and zelinxu and wenying.dai and Qingjie Liu and Yunhong Wang},
booktitle={The Fourteenth International Conference on Learning Representations},
year={2026}
}

@article{drivevla-w0,
  title={{DriveVLA-W0}: {World} Models Amplify Data Scaling Law in Autonomous Driving},
  author={Li, Yingyan and Shang, Shuyao and Liu, Weisong and Zhan, Bing and Wang, Haochen and Wang, Yuqi and Chen, Yuntao and Wang, Xiaoman and An, Yasong and Tang, Chufeng and others},
  journal={arXiv preprint arXiv:2510.12796},
  year={2025}
}

@article{law,
      title={Enhancing End-to-End Autonomous Driving with Latent World Model}, 
      author={Yingyan Li and Lue Fan and Jiawei He and Yuqi Wang and Yuntao Chen and Zhaoxiang Zhang and Tieniu Tan},
      year={2024},
      journal={arXiv preprint arXiv:2406.08481},
      archivePrefix={arXiv},
      primaryClass={cs.CV}
}

@article{dinotok,
  title={{DINO-Tok}: {Adapting} {DINO} for Visual Tokenizers}, 
  author={Mingkai Jia and Mingxiao Li and Zhijian Shu and Anlin Zheng and Liaoyuan Fan and Jiaxin Guo and Tianxing Shi and Dongyue Lu and Zeming Li and Xiaoyang Guo and Xiaojuan Qi and Xiao-Xiao Long and Qian Zhang and Ping Tan and Wei Yin},
  year={2026},
  journal={arXiv preprint arXiv:2511.20565},
  archivePrefix={arXiv},
  primaryClass={cs.CV}
}

@inproceedings{flow-ode,
title={Building Normalizing Flows with Stochastic Interpolants},
author={Michael Samuel Albergo and Eric Vanden-Eijnden},
booktitle={The Eleventh International Conference on Learning Representations },
year={2023}
}

@inproceedings{flow-matching,
title={Flow Matching for Generative Modeling},
author={Yaron Lipman and Ricky T. Q. Chen and Heli Ben-Hamu and Maximilian Nickel and Matthew Le},
booktitle={The Eleventh International Conference on Learning Representations },
year={2023}
}

@inproceedings{rectified-flow,
title={Flow Straight and Fast: {Learning} to Generate and Transfer Data with Rectified Flow},
author={Xingchao Liu and Chengyue Gong and qiang liu},
booktitle={The Eleventh International Conference on Learning Representations },
year={2023}
}

@inproceedings{navsim-v1,
	author = {Wei Cao and Marcel Hallgarten and Tianyu Li and Daniel Dauner and Xunjiang Gu and Caojun Wang and Yakov Miron and Marco Aiello and Hongyang Li and Igor Gilitschenski and Boris Ivanovic and Marco Pavone and Andreas Geiger and Kashyap Chitta}, 
	title = {Pseudo-Simulation for Autonomous Driving}, 
	booktitle = {Conference on Robot Learning (CoRL)}, 
	year = {2025}, 
}

@inproceedings{navsim-v2,
	title = {{NAVSIM}: {Data}-Driven Non-Reactive Autonomous Vehicle Simulation and Benchmarking},
	author = {Daniel Dauner and Marcel Hallgarten and Tianyu Li and Xinshuo Weng and Zhiyu Huang and Zetong Yang and Hongyang Li and Igor Gilitschenski and Boris Ivanovic and Marco Pavone and Andreas Geiger and Kashyap Chitta},
	booktitle = {Advances in Neural Information Processing Systems (NeurIPS)},
	year = {2024},
}

@article{svg-t2i,
      title={{SVG-T2I}: {Scaling} Up Text-to-Image Latent Diffusion Model Without Variational Autoencoder}, 
      author={Minglei Shi and Haolin Wang and Borui Zhang and Wenzhao Zheng and Bohan Zeng and Ziyang Yuan and Xiaoshi Wu and Yuanxing Zhang and Huan Yang and Xintao Wang and Pengfei Wan and Kun Gai and Jie Zhou and Jiwen Lu},
      year={2025},
      journal={arXiv preprint arXiv:2512.11749},
      archivePrefix={arXiv},
      primaryClass={cs.CV}
}

@article{svg,
      title={Latent Diffusion Model without Variational Autoencoder}, 
      author={Minglei Shi and Haolin Wang and Wenzhao Zheng and Ziyang Yuan and Xiaoshi Wu and Xintao Wang and Pengfei Wan and Jie Zhou and Jiwen Lu},
      year={2025},
      journal={arXiv preprint arXiv:2510.15301},
      archivePrefix={arXiv},
      primaryClass={cs.CV}
}

@inproceedings{pwm,
  title={{From Forecasting to Planning}: {Policy} World Model for Collaborative State-Action Prediction},
  author={Zhao, Zhida and Fu, Talas and Wang, Yifan and Wang, Lijun and Lu, Huchuan},
  booktitle={Advances in Neural Information Processing Systems},
  year={2025}
}

@inproceedings{drivelaw,
  title={Drivelaw: Unifying planning and video generation in a latent driving world},
  author={Xia, Tianze and Li, Yongkang and Zhou, Lijun and Yao, Jingfeng and Xiong, Kaixin and Sun, Haiyang and Wang, Bing and Ma, Kun and Chen, Guang and Ye, Hangjun and others},
  booktitle={Proceedings of the IEEE/CVF Conference on Computer Vision and Pattern Recognition},
  pages={39701--39712},
  year={2026}
}

@inproceedings{driveworld,
  title={Driveworld: 4d pre-trained scene understanding via world models for autonomous driving},
  author={Min, Chen and Zhao, Dawei and Xiao, Liang and Zhao, Jian and Xu, Xinli and Zhu, Zheng and Jin, Lei and Li, Jianshu and Guo, Yulan and Xing, Junliang and others},
  booktitle={Proceedings of the IEEE/CVF conference on computer vision and pattern recognition},
  pages={15522--15533},
  year={2024}
}

@inproceedings{drivewm,
  title={Driving into the future: Multiview visual forecasting and planning with world model for autonomous driving},
  author={Wang, Yuqi and He, Jiawei and Fan, Lue and Li, Hongxin and Chen, Yuntao and Zhang, Zhaoxiang},
  booktitle={Proceedings of the IEEE/CVF Conference on Computer Vision and Pattern Recognition},
  pages={14749--14759},
  year={2024}
}

@article{li2025imagidrive,
  title={ImagiDrive: A Unified Imagination-and-Planning Framework for Autonomous Driving},
  author={Li, Jingyu and Zhang, Bozhou and Jin, Xin and Deng, Jiankang and Zhu, Xiatian and Zhang, Li},
  journal={arXiv preprint arXiv:2508.11428},
  year={2025}
}

@article{yang2025worldrft,
  title={WorldRFT: Latent World Model Planning with Reinforcement Fine-Tuning for Autonomous Driving},
  author={Yang, Pengxuan and Lu, Ben and Xia, Zhongpu and Han, Chao and Gao, Yinfeng and Zhang, Teng and Zhan, Kun and Lang, XianPeng and Zheng, Yupeng and Zhang, Qichao},
  journal={arXiv preprint arXiv:2512.19133},
  year={2025}
}

@article{gui2026bridging,
  title={Bridging scene generation and planning: Driving with world model via unifying vision and motion representation},
  author={Gui, Xingtai and Zhang, Meijie and Yan, Tianyi and Han, Wencheng and Gong, Jiahao and Tan, Feiyang and Xu, Cheng-zhong and Shen, Jianbing},
  journal={arXiv preprint arXiv:2603.14948},
  year={2026}
}

@article{bardes2024v,
  title={Revisiting feature prediction for learning visual representations from video},
  author={Bardes, Adrien and Garrido, Quentin and Ponce, Jean and Chen, Xinlei and Rabbat, Michael and LeCun, Yann and Assran, Mahmoud and Ballas, Nicolas},
  journal={arXiv preprint arXiv:2404.08471},
  year={2024}
}

@article{yu2024representation,
  title={Representation alignment for generation: Training diffusion transformers is easier than you think},
  author={Yu, Sihyun and Kwak, Sangkyung and Jang, Huiwon and Jeong, Jongheon and Huang, Jonathan and Shin, Jinwoo and Xie, Saining},
  journal={arXiv preprint arXiv:2410.06940},
  year={2024}
}

@inproceedings{radford2021learning,
  title={Learning transferable visual models from natural language supervision},
  author={Radford, Alec and Kim, Jong Wook and Hallacy, Chris and Ramesh, Aditya and Goh, Gabriel and Agarwal, Sandhini and Sastry, Girish and Askell, Amanda and Mishkin, Pamela and Clark, Jack and others},
  booktitle={International conference on machine learning},
  pages={8748--8763},
  year={2021},
  organization={PmLR}
}

@article{maes2026leworldmodel,
  title={Leworldmodel: Stable end-to-end joint-embedding predictive architecture from pixels},
  author={Maes, Lucas and Lidec, Quentin Le and Scieur, Damien and LeCun, Yann and Balestriero, Randall},
  journal={arXiv preprint arXiv:2603.19312},
  year={2026}
}

@article{tu2025survey,
  title={The role of world models in shaping autonomous driving: A comprehensive survey},
  author={Tu, Sifan and Zhou, Xin and Liang, Dingkang and Jiang, Xingyu and Zhang, Yumeng and Li, Xiaofan and Bai, Xiang},
  journal={arXiv preprint arXiv:2502.10498},
  year={2025}
}

@inproceedings{drivingworld,
  title={Drivingworld: Constructing world model for autonomous driving via video gpt},
  author={Hu, Xiaotao and Jia, Mingkai and Guo, Xiaoyang and Zhang, Qian and Long, Xiao-xiao and Yin, Wei},
  booktitle={International Conference on Pattern Recognition},
  pages={276--291},
  year={2026},
  organization={Springer}
}

@article{eponav2,
  title={EponaV2: Driving World Model with Comprehensive Future Reasoning},
  author={Xu, Jiawei and Zhong, Zhizhou and Shu, Zhijian and Jia, Mingkai and Li, Mingxiao and Bian, Jia-Wang and Zhang, Qian and Zhang, Kaicheng and Xie, Jin and others},
  journal={arXiv preprint arXiv:2605.14696},
  year={2026}
}

@inproceedings{assran2023self,
  title={Self-supervised learning from images with a joint-embedding predictive architecture},
  author={Assran, Mahmoud and Duval, Quentin and Misra, Ishan and Bojanowski, Piotr and Vincent, Pascal and Rabbat, Michael and LeCun, Yann and Ballas, Nicolas},
  booktitle={2023 IEEE/CVF Conference on Computer Vision and Pattern Recognition (CVPR)},
  pages={15619--15629},
  year={2023},
  organization={IEEE}
}

@article{assran2025v,
  title={V-jepa 2: Self-supervised video models enable understanding, prediction and planning},
  author={Assran, Mido and Bardes, Adrien and Fan, David and Garrido, Quentin and Howes, Russell and Muckley, Matthew and Rizvi, Ammar and Roberts, Claire and Sinha, Koustuv and Zholus, Artem and others},
  journal={arXiv preprint arXiv:2506.09985},
  year={2025}
}

@article{wang2026drive,
  title={Drive-jepa: Video jepa meets multimodal trajectory distillation for end-to-end driving},
  author={Wang, Linhan and Yang, Zichong and Bai, Chen and Zhang, Guoxiang and Liu, Xiaotong and Zheng, Xiaoyin and Long, Xiao-Xiao and Lu, Chang-Tien and Lu, Cheng},
  journal={arXiv preprint arXiv:2601.22032},
  year={2026}
}

@article{yang2026auto,
  title={Auto-JEPA: A Latent World Model of Continuous Intent for End-to-End Autonomous Driving},
  author={Yang, Jiwei and Chen, Zhengxian and Huang, Chaosheng and Li, Jun},
  journal={arXiv preprint arXiv:2607.29031},
  year={2026}
}

@article{liu2026driveworldvla,
  title={Driveworld-vla: Unified latent-space world modeling with vision-language-action for autonomous driving},
  author={Liu, Lin and Song, Ziying and Jia, Caiyan and Ye, Hangjun and Hao, Xiaoshuai and Chen, Long and others},
  journal={arXiv preprint arXiv:2602.06521},
  year={2026}
}

@article{zhao2025autoregressive,
  title={Autoregressive End-to-End Planning with Time-Invariant Spatial Alignment and Multi-Objective Policy Refinement},
  author={Zhao, Jianbo and Ban, Taiyu and Li, Xiangjie and Gui, Xingtai and Zhou, Hangning and Liu, Lei and Zhao, Hongwei and Li, Bin},
  journal={arXiv preprint arXiv:2509.20938},
  year={2025}
}

@article{luo2025adathinkdrive,
  title={Adathinkdrive: Adaptive thinking via reinforcement learning for autonomous driving},
  author={Luo, Yuechen and Li, Fang and Xu, Shaoqing and Lai, Zhiyi and Yang, Lei and Chen, Qimao and Luo, Ziang and Xie, Zixun and others},
  journal={arXiv preprint arXiv:2509.13769},
  year={2025}
}

@article{xing2025mimir,
  title={Mimir: Hierarchical goal-driven diffusion with uncertainty propagation for end-to-end autonomous driving},
  author={Xing, Zebin and Zheng, Yupeng and Zhang, Qichao and Ding, Zhixing and Yang, Pengxuan and Gu, Songen and Xia, Zhongpu and Zhao, Dongbin},
  journal={IEEE Robotics and Automation Letters},
  volume={11},
  number={2},
  pages={2178--2185},
  year={2025},
  publisher={IEEE}
}

@article{wozniak2026prix,
  title={Prix: Learning to plan from raw pixels for end-to-end autonomous driving},
  author={Wozniak, Maciej and Liu, Lianhang and Cai, Yixi and Jensfelt, Patric},
  journal={IEEE Robotics and Automation Letters},
  volume={11},
  number={5},
  pages={6400--6407},
  year={2026},
  publisher={IEEE}
}

@article{feng2025artemis,
  title={Artemis: Autoregressive end-to-end trajectory planning with mixture of experts for autonomous driving},
  author={Feng, Renju and Xi, Ning and Chu, Duanfeng and Wang, Rukang and Deng, Zejian and Wang, Anzheng and Lu, Liping and Wang, Jinxiang and Huang, Yanjun},
  journal={IEEE Robotics and Automation Letters},
  volume={11},
  number={1},
  pages={226--233},
  year={2025},
  publisher={IEEE}
}

@article{zhong2025anytalker,
  title={Anytalker: Scaling multi-person talking video generation with interactivity refinement},
  author={Zhong, Zhizhou and Ji, Yicheng and Kong, Zhe and Liu, Yiying and Wang, Jiarui and Feng, Jiasun and Liu, Lupeng and Wang, Xiangyi and Li, Yanjia and She, Yuqing and others},
  journal={arXiv preprint arXiv:2511.23475},
  year={2025}
}

@inproceedings{mi2025data,
  title={Data synthesis with diverse styles for face recognition via 3dmm-guided diffusion},
  author={Mi, Yuxi and Zhong, Zhizhou and Huang, Yuge and Yuan, Qiuyang and Zhao, Xuan and Xu, Jianqing and Ding, Shouhong and Wang, Shaoming and Guo, Rizen and Zhou, Shuigeng},
  booktitle={2025 IEEE/CVF Conference on Computer Vision and Pattern Recognition (CVPR)},
  pages={21203--21214},
  year={2025},
  organization={IEEE}
}

@inproceedings{chen2025focused,
  title={A focused human body model for accurate anthropometric measurements extraction},
  author={Chen, Shuhang and Huang, Xianliang and Zhong, Zhizhou and Guan, Juhong and Zhou, Shuigeng},
  booktitle={2025 IEEE/CVF Conference on Computer Vision and Pattern Recognition (CVPR)},
  pages={22658--22667},
  year={2025},
  organization={IEEE}
}

@article{guo2026salon3r,
  title={SaLon3R: Structure-Aware Long-Term Feedforward 3D Reconstruction from Unposed Images},
  author={Guo, Jiaxin and Guan, Tongfan and Dong, Wenzhen and Zheng, Wenzhao and Wang, Wenting and Wang, Yue and Yam, Yeung and Liu, Yun-Hui},
  journal={International Journal of Computer Vision},
  volume={134},
  number={8},
  pages={383},
  year={2026},
  publisher={Springer}
}

@inproceedings{xu2025ad,
  title={Ad-gs: Object-aware b-spline gaussian splatting for self-supervised autonomous driving},
  author={Xu, Jiawei and Deng, Kai and Fan, Zexin and Wang, Shenlong and Xie, Jin and Yang, Jian},
  booktitle={2025 IEEE/CVF International Conference on Computer Vision (ICCV)},
  pages={24770--24779},
  year={2025},
  organization={IEEE}
}

@inproceedings{guo2025endo3r,
  title={Endo3r: Unified online reconstruction from dynamic monocular endoscopic video},
  author={Guo, Jiaxin and Dong, Wenzhen and Huang, Tianyu and Ding, Hao and Wang, Ziyi and Kuang, Haomin and Dou, Qi and Liu, Yun-Hui},
  booktitle={International Conference on Medical Image Computing and Computer-Assisted Intervention},
  pages={170--180},
  year={2025},
  organization={Springer}
}

@inproceedings{guo2022visual,
  title={A visual navigation perspective for category-level object pose estimation},
  author={Guo, Jiaxin and Zhong, Fangxun and Xiong, Rong and Liu, Yunhui and Wang, Yue and Liao, Yiyi},
  booktitle={European Conference on Computer Vision},
  pages={123--141},
  year={2022},
  organization={Springer}
}

@article{jia2025mgvq,
  title={Mgvq: Could vq-vae beat vae? a generalizable tokenizer with multi-group quantization},
  author={Jia, Mingkai and Yin, Wei and Hu, Xiaotao and Guo, Jiaxin and Guo, Xiaoyang and Zhang, Qian and Long, Xiao-Xiao and Tan, Ping},
  journal={arXiv preprint arXiv:2507.07997},
  year={2025}
}

@article{kong20253d,
  title={3d and 4d world modeling: A survey},
  author={Kong, Lingdong and Yang, Yu and Mei, Jianbiao and Liu, Youquan and Liang, Ao and Zhu, Dekai and Lu, Dongyue and Yin, Wei and Hu, Xiaotao and Jia, Mingkai and others},
  journal={arXiv preprint arXiv:2509.07996},
  year={2025}
}

@inproceedings{cheng2024mf,
  title={Mf-mos: A motion-focused model for moving object segmentation},
  author={Cheng, Jintao and Zeng, Kang and Huang, Zhuoxu and Tang, Xiaoyu and Wu, Jin and Zhang, Chengxi and Chen, Xieyuanli and Fan, Rui},
  booktitle={2024 IEEE International Conference on Robotics and Automation (ICRA)},
  pages={12499--12505},
  year={2024},
  organization={IEEE}
}

@article{zhou2026thinklocallyrefineglobally,
      title={Think Locally, Refine Globally for Memory-Efficient 3D Reconstruction}, 
      author={Jingke Zhou and Chenhang Ma and Zhizhou Zhong and Mingkai Liu and Zhuang Zhou and Yicheng ji and Binghua Su and Bo Cai and Xianliang Huang},
      journal={arXiv preprint arXiv:2609.21437},
      year={2026}
}
\end{document}